\documentclass[letterpaper, 10 pt, conference]{ieeeconf}  
\IEEEoverridecommandlockouts

\usepackage[utf8]{inputenc} 
\usepackage[T1]{fontenc}
\usepackage{amsmath}
\usepackage{amsfonts}
\usepackage{float}
\usepackage{graphicx}
\usepackage{booktabs}
\usepackage[font=small]{caption}
\usepackage{subcaption}
\usepackage[ruled]{algorithm2e}
\usepackage{booktabs} 
\usepackage{enumerate}
\usepackage{times}
\usepackage{soul}
\usepackage{url}
\usepackage{hyperref}
\usepackage{bm}
\usepackage{adjustbox}
\usepackage{xcolor}
\usepackage{placeins}
\usepackage[utf8]{inputenc}
\usepackage{booktabs}
\usepackage{array, multirow}
\usepackage[capitalise]{cleveref}
\usepackage{tikz}
\usepackage{wrapfig}
\usepackage{adjustbox}
\usepackage{multirow}
\usepackage{amssymb}
\usepackage{pifont}
\usepackage{algorithmic}
\usepackage{cleveref}
\usepackage{mathabx}
\newcommand{\bb}{\mathbf}  
\newcommand{\R}{\mathbf{R}}
\newcommand{\N}{\mathcal{N}}

\newcommand{\SE}{\mathbf{SE}}

\begin{document}

\newcommand\mycommfont[1]{\footnotesize\rmfamily\textcolor{blue}{#1}}
\usetikzlibrary{arrows.meta}
\usetikzlibrary{positioning}
\tikzstyle{decision} = [diamond, draw, fill=blue!20, 
    text width=6em, text badly centered, node distance=3cm, inner sep=0pt]
\tikzstyle{block} = [rectangle, draw, fill=gray!10, 
    text width=10em, very thick, text centered, rounded corners, minimum height=2.2em]
\tikzstyle{line} = [draw, -{latex[scale=15.0]}]
\tikzstyle{cloud} = [draw, ellipse,fill=red!20, node distance=3cm,
    minimum height=2em]
\setlength{\fboxrule}{1pt}
\setlength{\fboxsep}{0pt}  

\newcounter{task}
\newcommand{\task}[2]{%
  \refstepcounter{task}%
  \label{#1}%
  \textit{Task~\thetask\ -- #2}%
}

\newcommand{\red}[1]{\textcolor{black}{#1}}
\newcommand{\green}[1]{\textcolor{green}{#1}}
\newcommand{\blue}[1]{\textcolor{blue}{#1}}

\captionsetup{skip=1pt}
\setlength{\textfloatsep}{1pt}
\setlength{\belowdisplayskip}{1pt} \setlength{\belowdisplayshortskip}{1pt}
\setlength{\abovedisplayskip}{1pt} \setlength{\abovedisplayshortskip}{1pt}
\setlength{\floatsep}{1pt} \setlength{\textfloatsep}{1pt}
\setlength{\intextsep}{1pt}
\setlength{\abovecaptionskip}{1pt}
\setlength{\belowcaptionskip}{1pt}
\captionsetup{belowskip=1pt}

\newcommand{\cmark}{\ding{51}}%
\newcommand{\xmark}{\ding{55}}%

\newtheorem{innercustomthm}{Theorem}
\newenvironment{customthm}[1]
  {\renewcommand\theinnercustomthm{#1}\innercustomthm}
  {\endinnercustomthm}

\newtheorem{innercustomprop}{Proposition}
\newenvironment{customprop}[1]
  {\renewcommand\theinnercustomprop{#1}\innercustomprop}
  {\endinnercustomprop}

\newtheorem{definition}{Definition}[section]
\newtheorem{prop}{Proposition}[section]
\newtheorem{theorem}{Theorem}[section]
  
\captionsetup[figure]{size=small}

\title{\LARGE \bf Joint Flow Trajectory Optimization For Feasible\\ Robot Motion Generation from Video Demonstrations
}

\author{Xiaoxiang Dong$^{1,2}$ \and Matthew Johnson-Roberson$^{1,2}$ \and Weiming Zhi$^{1,2,3,*}$
\thanks{$^{*}$email: {\tt\small Weiming.Zhi@sydney.edu.au}.}%
\thanks{$^{1}$ College of Connected Computing, Vanderbilt University, Nashville, TN, USA}
\thanks{$^{2}$ Robotics Institute, Carnegie Mellon University, Pittsburgh, PA, USA}
\thanks{$^{3}$ School of Computer Science, The University of Sydney, Australia.}
%
}
\maketitle

\begin{abstract}
Learning from human video demonstrations offers a scalable alternative to teleoperation or kinesthetic teaching, but poses challenges for robot manipulators due to embodiment differences and joint feasibility constraints. We address this problem by proposing the Joint Flow Trajectory Optimization (JFTO) framework for grasp pose generation and object trajectory imitation under the video-based Learning-from-Demonstration (LfD) paradigm. Rather than directly imitating human hand motions, our method treats demonstrations as object-centric guides, balancing three objectives: (i) selecting a feasible grasp pose, (ii) generating object trajectories consistent with demonstrated motions, and (iii) ensuring collision-free execution within robot kinematics. To capture the multimodal nature of demonstrations, we extend flow matching to $\SE(3)$ for probabilistic modeling of object trajectories, enabling density-aware imitation that avoids mode collapse. The resulting optimization integrates grasp similarity, trajectory likelihood, and collision penalties into a unified differentiable objective. We validate our approach in both simulation and real-world experiments across diverse real-world manipulation tasks. 
\end{abstract}


\section{Introduction}
Generating feasible motions is a key challenge in robotics. Traditional approaches address this by formulating and solving optimization problems with carefully designed costs and constraints \cite{rrts,GeoFab_gloabL_opt, chomp}. An alternative line of work focuses on enabling robots to imitate humans through \emph{Learning from Demonstration} (LfD) \cite{ravichandar2020recent, Diff_templates}. Historically, this has required collecting demonstrations through teleoperation or precise kinesthetic teaching. More recently, researchers have attempted to leverage videos of human interaction with objects as demonstrations \cite{zhou2025you, bahety2025safemimic}. Although promising, this paradigm introduces challenges: Human arms and robot manipulators differ significantly in morphology, and direct imitation can cause robots to violate joint constraints when tracking human motions. All demonstrations in this paper refer to such human video demonstrations.

\begin{figure}[t]
    \centering
    \includegraphics[width=\columnwidth]{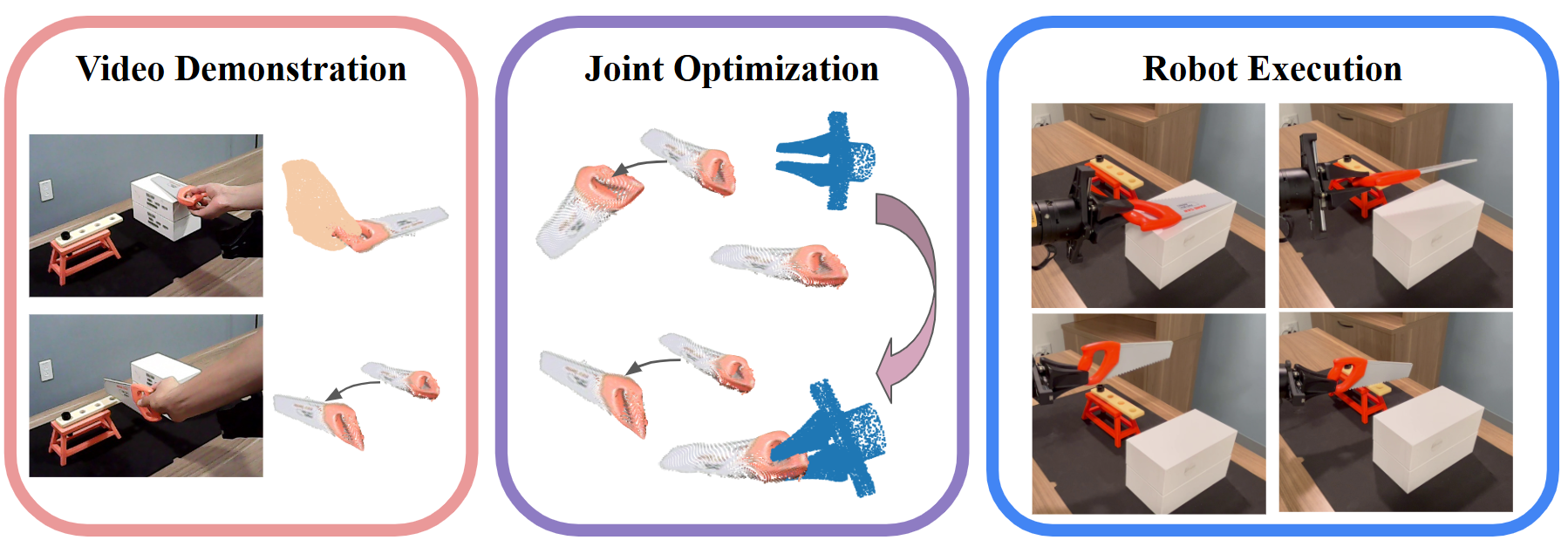}
    \caption{We address generating feasible grasp poses and motion trajectories from video-based LfD, ensuring that robot grasps and object motions remain consistent with human demonstrations despite constraints. We propose the Joint Flow Trajectory Optimization (JFTO) framework, which jointly optimizes over grasp feasibility and trajectory imitation, producing a gripper pose and motion plan that closely match the demonstration while remaining executable.}
    \label{fig:figure_1}
\end{figure}

In this paper, we tackle the problem of generating grasp poses and motion trajectories for robot manipulators under the video-based LfD paradigm, where the tracked human motion may often not be kinesthetically feasible for a robot manipulator to execute. Our goal is to identify feasible grasp configurations and ensure that grasped objects move consistently with human video demonstrations. Concretely, given videos of humans interacting with objects, we propose the Joint Flow Trajectory Optimization (JFTO) framework, as shown in \cref{fig:figure_1}, with three desiderata: (1) identifying a feasible grasp pose; (2) generating trajectories of the grasped object that align with object poses in the videos; (3) ensuring downstream robot motions remain collision-free and within joint limits. To this end, our formulation treats human video demonstrations as object-centric guides rather than strict motion references. By focusing on how objects are manipulated, rather than how the human hand is configured, we allow robots to synthesize grasps and trajectories that are consistent with observed outcomes, while respecting the robot’s own embodiment.
Uncertainty is critical to robotics tasks \cite{senanayake2024role, constr, vasudevan2025strategic}. We take a probabilistic approach and model the distribution of object pose trajectories extracted from videos as a generative model \emph{Flow Matching} \cite{lipman2022flow} over the Special Euclidean Group $\bb{SE}(3)$.
Here, as the grasps and motions of the human from the video-based demonstrations may not be valid for robot execution, our goal is not only to ensure grasp and execution feasibility, but also to mimic the motion patterns of human demonstrations in the videos. 

Concretely, the technical contributions of this work are threefold:
\begin{itemize}
    \item We introduce the Joint Flow Trajectory Optimization (JFTO) framework that integrates grasp similarity, object trajectory likelihood, and collision avoidance into a unified differentiable objective for video-based LfD.
    \item We extend flow matching to $\SE(3)$ for probabilistic modeling of demonstrated object trajectories, enabling density-aware imitation that preserves multi-modality.
    \item We demonstrate the effectiveness of JFTO in both simulation and real-world experiments on diverse manipulation tasks, showing that joint optimization achieves higher fidelity to demonstrations compared to sequential baselines.
\end{itemize}

\section{Related Work}\label{sec:related_work}
This paper addresses joint generation of grasp poses and motion trajectories from video-based LfD, ensuring manipulator feasibility while remaining consistent with human demonstrations, relating to work in \emph{robot learning from videos} and \emph{grasp pose generation}.

\textbf{Robot Learning from Videos:} Human videos naturally capture interaction, motion, and affordances, providing rich supervision for robot learning. Recent work leverages such data to build transferable visuomotor skills, from single demonstrations to large-scale web video distillation. For example, \emph{Human2Sim2Robot} \cite{lum2025crossinghumanrobotembodimentgap} mitigates the human--robot embodiment gap by extracting object pose trajectories and premanipulation hand configurations from a single RGB-D video, using them to define object-centric rewards and initialize reinforcement learning (RL) in simulation, enabling zero-shot sim-to-real transfer of dexterous policies. At the other end, \emph{ZeroMimic} \cite{shi2025zeromimic} distills manipulation skills from large-scale in-the-wild datasets such as EpicKitchens, producing image goal-conditioned policies for tasks like opening, pouring, and stirring that generalize across embodiments. Other approaches emphasize one-shot or safety-aware imitation \cite{zhou2025you,bahety2025safemimic,kareer2024egomimic}, or synthetic demonstrations for robustness and efficiency \cite{huang2024rekep,chen2025tool}. Earlier work also explored learning from unstructured human videos via scalable datasets \cite{bahl2022human,herd:liu:2022} or embodied dexterous platforms \cite{shaw2024demonstrating}. Overall, relative to other interfaces for demonstration collection \cite{zhi2023learning}, learning from videos allows for larger scale data collection.

\textbf{Grasp Pose Generation:} The Grasp Pose Generator (GPG), first introduced in the Grasp Pose Detection framework \cite{tenpas2017gpd}, samples candidate grasps from point cloud geometry. Candidates are filtered with heuristics (antipodal checks, collision tests) before evaluation by analytic or learned models. GPG supports large-scale datasets by labeling analytic positives (antipodal, force closure) and negatives, a paradigm popularized by Gualtieri et al.~\cite{gualtieri2016gpd} and widely adopted in training pipelines \cite{liang2019pointnetgpd,sundermeyer2021contactgraspnet}. Methods which explore different object representations \cite{RecGS} upstream to grasping have been explored \cite{IchnowskiAvigal2021DexNeRF}. Benchmarks like \emph{GraspNet-1Billion} \cite{fang2020graspnet} follow this approach, providing more than a billion simulated grasps. Recent refinements, such as \emph{Graspness Discovery} \cite{wang2024graspness}, introduce geometric quality measures to faster pruning and higher sample quality. GPG also serves as a sampler in modern data-driven pipelines. \emph{GtG 2.0} \cite{rashidi2025gtg2} combines GPG with graph neural networks to score 7-DoF grasps, boosting precision on GraspNet-1Billion by 35\% and achieving over 90\% success in real trials. Likewise, \emph{DexGraspNet 2.0} \cite{zhang2024dexgraspnet2} generates millions of dexterous grasps in simulation, validated with physics checks before training generative models.

\section{Preliminaries: Flow Matching}\label{sec:preliminaries}
Flow matching is a generative modeling approach that estimates continuous data distributions by learning vector fields that transform samples from a known prior distribution to a target distribution by integration along an Ordinary Differential Equation (ODE). This method provides a fast and exact generative process and avoids many of the complexities associated with likelihood-based models or score-based models \cite{chen2023otfm, GeoFab_gloabL_opt, albergo2023stochastic}. Flow matching speeds up previous normalizing flow methods \cite{normalising_flow} and their applications in robotics \cite{periodic,Fast_diff_int}.

\begin{figure}[t]
    \centering
    \includegraphics[width=0.9\linewidth]{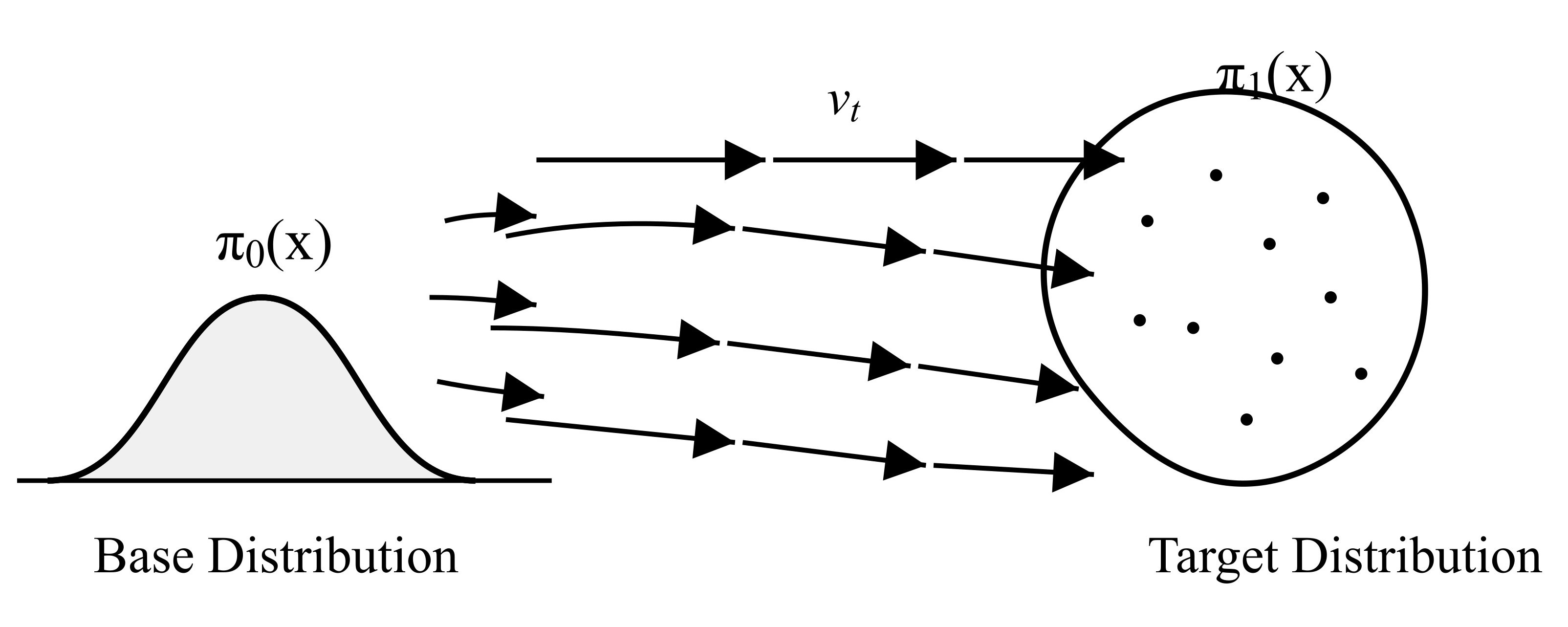}
    
    \includegraphics[width=0.9\linewidth]{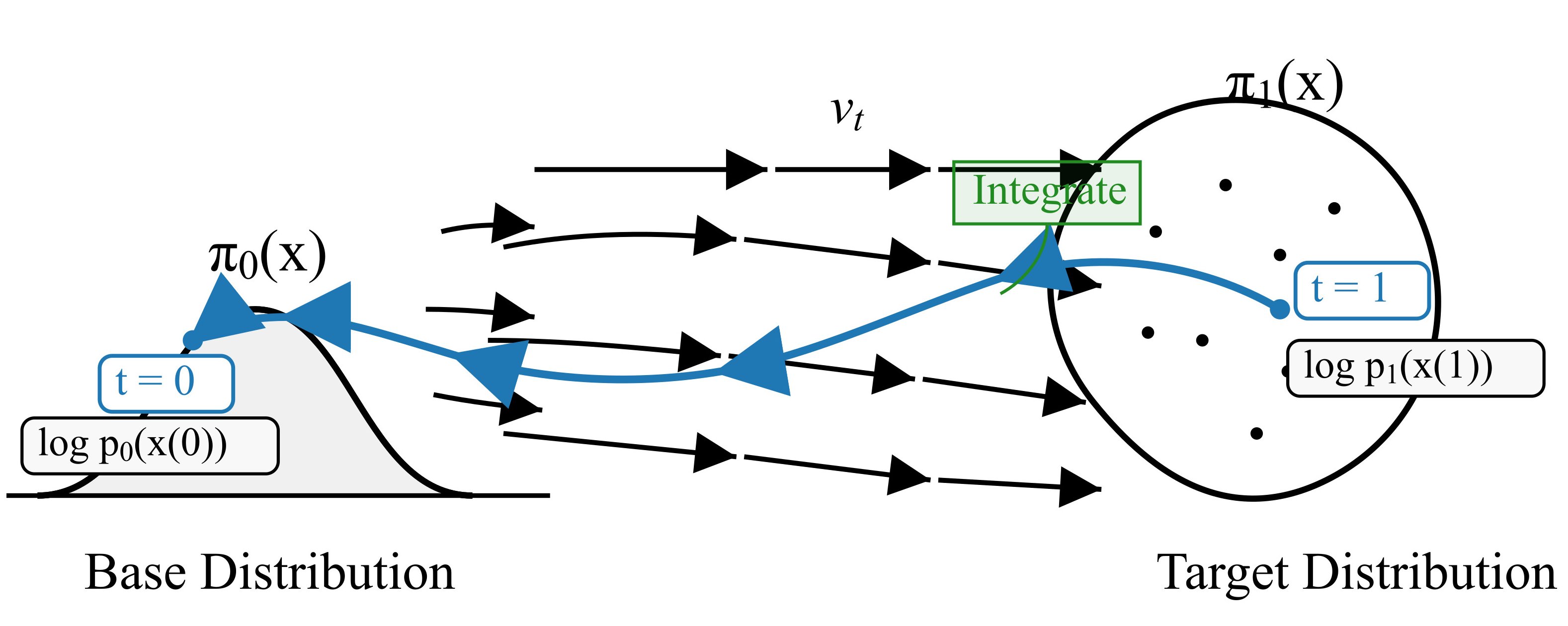}
    \caption{Flow Matching learns a time-dependent velocity field $v_t$ from based distribution $\pi_0(x)$ to target distribution $\pi_1(x)$, while integrating backward from a sample in $\pi_1(x)$ through $v_t$ to $\pi_0(x)$ derives the log-density of the sample.}
    \label{fig:flow_matching}
\end{figure}

Let $\pi_0(x)$ denote a base distribution over $x \in \R$, for example standard Gaussian $\N(0, 1)$, and let $\pi_1(x)$ denote the target distribution. Flow matching seeks to learn a time-dependent vector field $v_t: \ \R^{d} \rightarrow \R^{d}$ that defines a flow via the ODE: 
\begin{align}
    \frac{\bb{dx}}{\bb{dt}} = v_t(\bb{x}), && 
    \bb{x_0} \sim \bb{\pi_0},
\end{align}
such that the solution $\bb{x(1)}$ approximates the distribution $pi_1$, as shown in \cref{fig:flow_matching}.

To train this vector field, one sample $(\bb{x_0}, \bb{x_1}) \sim (\pi_0, \pi_1)$ and defines an interpolation:
\begin{align}
    \bb{x_t} = \gamma(t; \bb{x_0}, \bb{x_1}),
\end{align}
The target velocity field $\hat{v_t}(\bb{x_t})$ is then defined as the time derivative of the interpolation curve:
\begin{align}
    \hat{v_t}(\bb{x_t}) = \frac{d}{dt} \gamma(t; \bb{x_0}, \bb{x_1}),
\end{align}
The model is trained by minimizing the expected squared error between the learned field $v_t(\bb{x_t})$ and the target field $\dot{v_t}(\bb{x_t})$:
\begin{align}
    \mathbb{E}_{t \sim \mathcal{U}(0, 1)} \left[ \left\| v_t(\bb{x_t}) - \hat{v_t}(\bb{x_t}) \right\|^2 \right],
\end{align}
Once trained, the learned vector field can be used in two ways:
$1)$ generate new samples from $\pi_1$ by integrating the ODE from $\bb{x_0}$ to $\bb{x_1}$, 
$2)$ evaluate the log-density of a sample x(t) evolving under the flow using the instantaneous change of variables formula:
\begin{align}
    \frac{d}{dt} \log p_t(x(t)) = - \nabla \cdot v_t(x(t)),
\end{align}
which, when integrated backward from $t=1$ to $t=0$, yields:
\begin{align}
    \log p_1(x(1)) = \log p_0(x(0)) - \int_0^1 \nabla \cdot v_t(x(t)) ~dt~,\label{eqn:log_density}
\end{align}
as illustrated in \cref{fig:flow_matching}. In \cref{sec:METHOD}, we represent the probabilities in the motion trajectory distributions with flow models.

\section{Joint Flow Trajectory Optimization}\label{sec:METHOD}

\begin{figure}[t]
    \centering
    \includegraphics[width=\columnwidth]{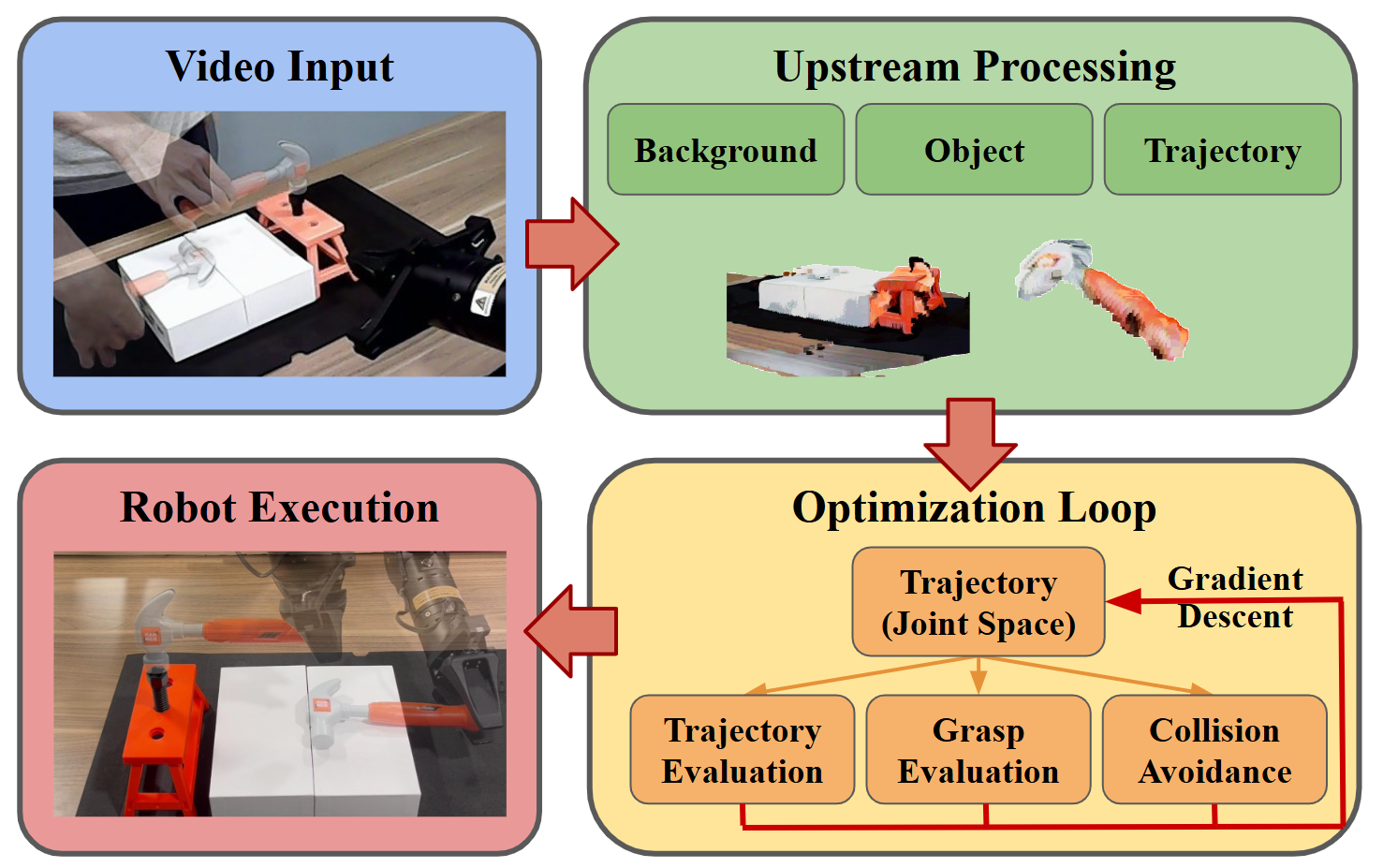}
    \caption{We present an overview of the Joint Flow Trajectory Optimization (JFTO) framework. Human demonstration videos are processed to extract the grasp pose and object trajectory. These are then used to jointly optimize a feasible execution trajectory in the robot’s joint space, and then executed on the robot.}
    \label{fig:methodology_pipeline}
\end{figure}

Here, we propose Joint Flow Trajectory Optimization (JFTO), a framework that \emph{jointly} optimizes the robot’s initial grasp and its subsequent motion so that the resulting object trajectory closely imitates a human video demonstration, while still remaining kinematically feasible and collision-free. Unlike sequential approaches that first fix a grasp and then attempt to optimize the trajectory, the JFTO formulation reasons about the grasp and downstream trajectory simultaneously. This allows the robot to choose grasps that remain feasible throughout the entire motion. An overview of JFTO is shown in \cref{fig:methodology_pipeline}. 

\subsection{Problem Formulation}

Let $\xi = \{q_t\}_{t=0}^{T}$ denote the joint states of the robot in the $T$ time steps. A rollout is scored by three components:
\begin{itemize}
    \item $S_T(\xi)$: how well the induced object trajectory matches the demonstrated trajectory,
    \item $S_G(q_0)$: how well the initial grasp resembles a feasible human-like grasp, and
    \item $S_C(\xi)$: how safe the rollout is with respect to joint limits and collisions.
\end{itemize}
The optimization problem is formulated as:
\begin{align}
    q^\star = \arg\max_{\xi} \Big[\alpha\, S_T(\xi) + \beta\, S_G(q_0) + \gamma\, S_C(\xi)\Big],\label{eqn:problem-formulation}
\end{align}
where $\alpha,\beta,\gamma$ are weights that balance the importance of trajectory imitation, grasp quality, and safety. Note that $q_0$ is a component of $\xi$.

This formulation emphasizes that the goal is not to directly replicate human hand motions, which may be kinematically infeasible, but rather to optimize for a joint trajectory that reproduces the demonstrated object motion and respects the constraints of the robot’s embodiment.

\subsection{Data Collection and Processing}\label{sec:DATA_COLC}

The proposed optimization framework assumes access to object poses obtained from human–object interactions. To this end, we introduce a system that extracts object trajectories directly from raw demonstration videos. While we focus on this video-based pipeline, the framework is agnostic to the source of demonstrations, and alternative methods for acquiring object pose information could be substituted without loss of generality.

For each task, we record synchronized image sequences of a human demonstration using two calibrated cameras. Each frame pair is passed to a 3D foundation model, in our case DUSt3R~\cite{Wang_2024_CVPR}, to reconstruct a dense 3D point cloud of the scene. Alternative models, such as MONSt3R \cite{zhang2024monst3r} can also be used. Since DUSt3R operates on an arbitrary internal scale, outlined in \cite{JCR, Sim_pose}, we recover the metric scale by triangulating the corresponding feature points between the image pairs and aligning their reconstructed 3D positions with the known physical camera baseline. This produces trajectories expressed in real-world coordinates.

\begin{figure}[t]
    \centering
    \includegraphics[width=\columnwidth]{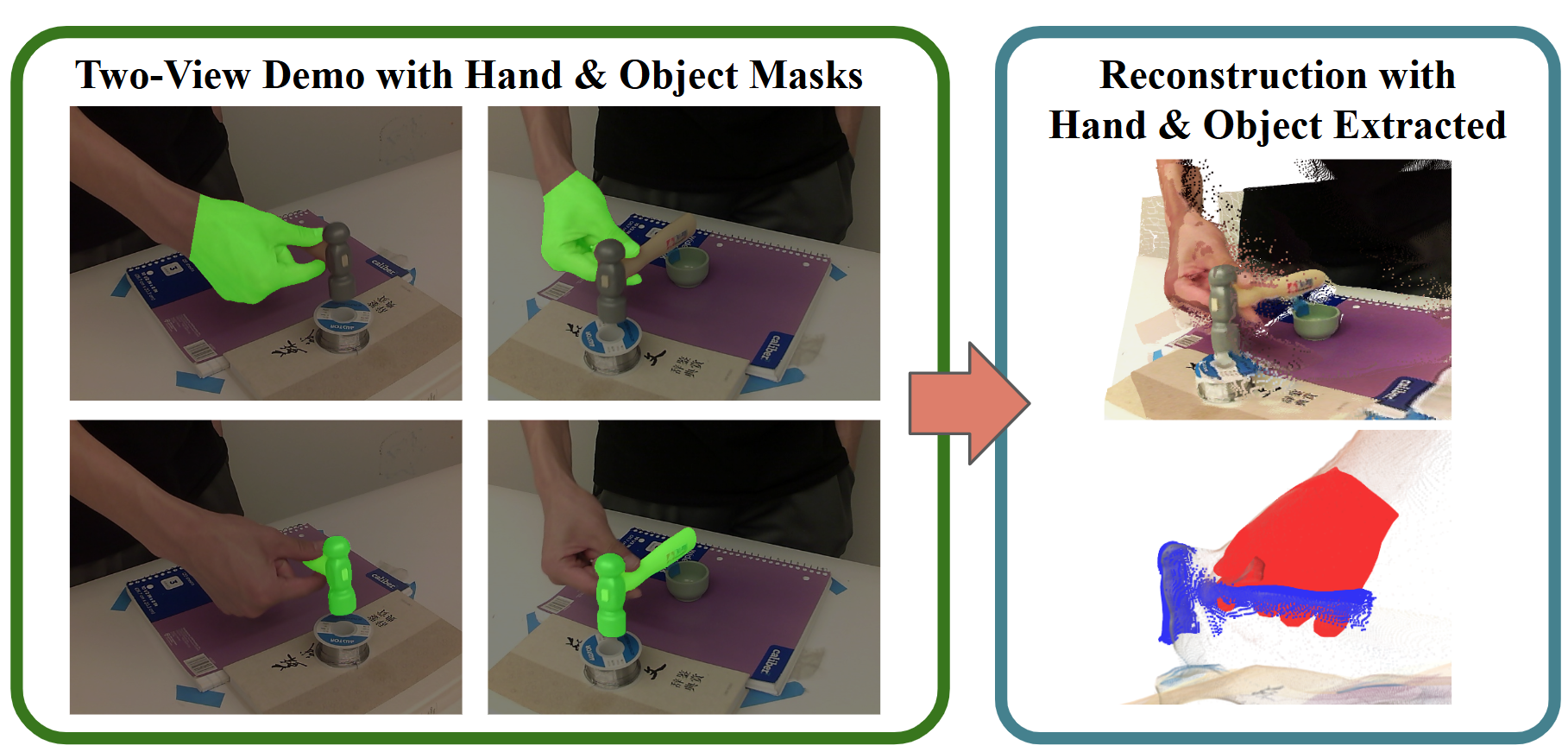}
    \caption{From two views of images with hand and object segmentation, we reconstruct the scene with 3D foundation models and use the segmentation masks to determine points corresponding to hand and object.}
    \label{fig:data-collection-demo}
\end{figure}

\textbf{Upstream Processing:} To isolate the relevant elements of the scene, we apply SAM2~\cite{ravi2024sam2} to segment both the human hand and the manipulated object in each frame. The segmentation masks allow us to separate the object and hand point clouds from the static background, as illustrated in \cref{fig:data-collection-demo}. We further estimate precise hand poses with HaMeR~\cite{pavlakos2024reconstructing}, repeating this process across key frames to obtain temporally consistent object–hand trajectories. An example visualization is shown in \cref{fig:trajectory-demo-vis}. To continuously track object motion, consecutive object point clouds are aligned using Iterative Closest Point (ICP) \cite{ICP_algo}, resulting in a sequence of estimated object poses $\{\hat{x}_t\}$ that describe its trajectory during the demonstration. In parallel, the background points across all frames are fused into a single static point cloud, which is later converted into a distance field for efficient collision checking.

\textbf{Connection to Downstream Optimization:} From each demonstration we extract the following structured outputs: (1) the object trajectory $\{\hat{x}_t\}$, which drives trajectory imitation ($S_T$); (2) the initial object pose $\hat{x}_0$, which anchors grasp evaluation ($S_G$); and (3) a static background model, which supports collision scoring ($S_C$). These components enable the downstream optimization of grasp and trajectory in a differentiable manner.

\subsection{Trajectory Quality Evaluation}\label{sec:TRAJ_METHOD}

\textbf{Flow Matching as Pose Likelihood Estimation:} From $n$ human demonstrations, each trajectory of $T$ waypoints in $\SE(3)$, we collect pose samples $\mathcal{Q}_t = \{x_t^{(i)}\}_{i=1}^n$ at every time step $t$. We treat $\mathcal{Q}_t$ as drawn from an underlying distribution $p_t(x)$ over $\SE(3)$. Our goal is to construct a density estimator
\[
f_{den}: \SE(3)\times \{0,\dots,T\} \rightarrow \mathbb{R},
\]
such that $f_{den}(x,t) \approx \log p_t(x)$ provides the log-likelihood of a candidate pose $x$ at time $t$.

\begin{figure}[t]
    \centering
    \includegraphics[width=\linewidth]{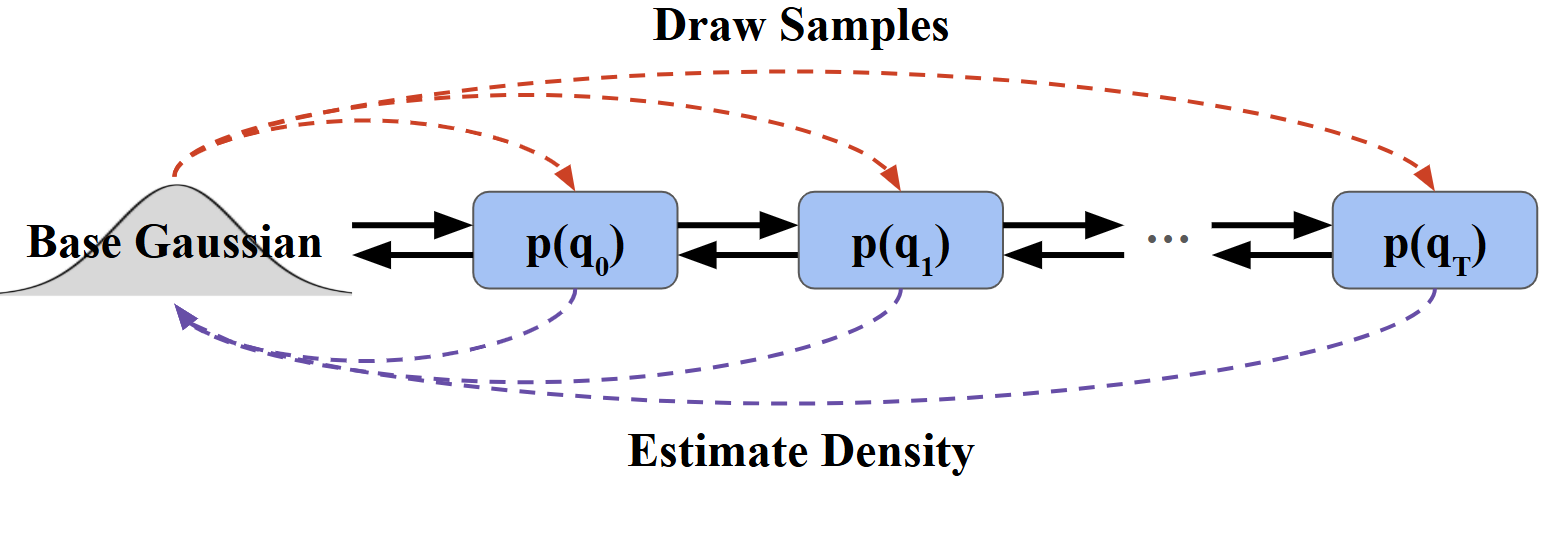}
    \caption{We treat each time step of the demonstration as a distribution over object poses in $\SE(3)$. Flow matching learns velocity fields that transform a base Gaussian into these distributions. This enables both sampling of realistic poses (forward integration) and computation of their log-likelihoods via divergence.}
    \label{fig:trajectory_flow}
\end{figure}

We use \emph{flow matching} to model these time-dependent pose distributions, as illustrated in \cref{fig:trajectory_flow}. Let $v_t:\SE(3)\rightarrow \mathbb{R}^6$ denote a learned velocity field, where the 6D output corresponds to the translational and rotational directions in the tangent space of $\SE(3)$. The field defines an ODE
\begin{align}
    \frac{d x_\tau}{d \tau} = v_\tau(x_\tau), \qquad x_0 \sim \pi_0,
\end{align}
where $\pi_0$ is a base Gaussian distribution over poses. Integrating this ODE from $\tau=0$ to $\tau=1$ pushes forward $\pi_0$ to approximate the demonstration distribution $p_{\tau}(x)$ at each step. The associated log-density can be computed using the instantaneous change-of-variables formula,
\begin{align}
    \log p_{\tau}(x) = \log \pi_0(x_0) - \int_0^t \nabla \cdot v_\tau(x_\tau)\, d\tau.
\end{align}
Thus, the density estimator is defined as
\begin{align}
    f_{den}(x) =\exp(\log p_{\tau}(x)),
\end{align}
which is tractable because both the ODE and its divergence are differentiable. Intuitively, $f_{den}(x)$ assigns high scores to poses that align with the demonstrated motion distribution, and low scores to outliers.

\textbf{Trajectory Rollout Evaluation:} To evaluate how well a candidate joint trajectory imitates the demonstrations, we begin with a sequence of joint states $\xi = \{q_t\}_{t=0}^{T}$ in the $d$-DoF robot joint space $\mathcal{J} \subset \mathbb{R}^d$. Each state is restricted to be within joint limits and is mapped through forward kinematics $F$ to obtain the corresponding pose of the end effector $r(q)_t \in \SE(3)$. This ensures that the trajectories are always kinematically feasible for execution. Given the initial pose of the object $\hat{x}_0$ recovered from the video, the relative transform between the gripper and the object at the first timestep is defined as 
\begin{align}
T_{go} = \hat{x}_0^{-1}\, r(q_0). 
\end{align}
This transform anchors the grasp and is assumed to be fixed throughout execution, meaning that subsequent gripper motions directly induce object motions. The resulting object trajectory is then expressed as $x(q_0) = r(q_0) \, T_{go}^{-1}$ for $t=0,\dots,T$. Each induced object pose is scored using the density estimator $f_{den}(x(q_t))$, and the rollout receives an overall trajectory score equal to the average log-likelihood across all time steps:
\begin{align}
    S_T(q) = \frac{1}{T}\sum_{t=0}^{T} f_{den}(x(q_t)).
\end{align}
In effect, this procedure asks whether the object trajectory implied by the robot motion falls within regions of high probability according to human demonstrations. Rollouts that closely reproduce demonstrated motion patterns earn high scores, while those that deviate substantially are penalized.

Intuitively, the density estimator $f_{den}$ models which poses are likely according to demonstrations. A rollout receives a high score if its induced object trajectory lies in regions of high demonstration density, thereby imitating not just the endpoints but the characteristic motion itself.

\subsection{Grasp Feasibility and Similarity Evaluations} \label{sec:GRASP_METHOD}

\subsubsection{Grasp Feasibility Check}
Successful execution of the object trajectory requires the gripper to establish and maintain a physically feasible grasp. To model grasp feasibility, we adopt a two-stage approach: (i) generate candidate positive grasps using a geometry-driven sampler, and (ii) train a learned classifier to distinguish these from infeasible alternatives. We first employ a Grasp Pose Generator (GPG)~\cite{tenpas2017gpd} to propose a set of analytically feasible grasps $P = \{p_i\}_{i=1}^N$ around the object point cloud. Each $p_i \in \SE(3)$ is guaranteed to satisfy local geometric conditions such as antipodal contact or force-closure heuristics and thus serves as a positive example.  

To train a discriminative model, we construct two types of negative examples:
\begin{itemize}
    \item \emph{Hard negatives}: generated by perturbing valid grasps with small translational and rotational noise, e.g.,
    \begin{align}
        p^-_{\text{hard}} = p_i \cdot \exp(\epsilon), \quad \epsilon \sim \mathcal{N}(0, \Sigma_{\text{small}}),
    \end{align}
    where $\exp(\cdot)$ maps a perturbation in the Lie algebra $\mathfrak{se}(3)$ to $\SE(3)$. These poses are close to feasible grasps, but violate stability constraints.
    \item \emph{Soft negatives}: sampled uniformly from the vicinity of the object in $\SE(3)$, without regard to contact conditions. These provide a broad coverage of infeasible grasps:
    \begin{align}
        p^-_{\text{soft}} \sim \mathcal{U}(\mathcal{N}_{\text{object}}) \subset \SE(3).
    \end{align}
\end{itemize}

Each pose $p \in \SE(3)$ is then embedded in a feature vector using a positional encoding $E:\SE(3)\rightarrow\mathbb{R}^e$. Specifically, the pose is first represented in a 6D form $(x,y,z,R_x,R_y,R_z)$, where $(x,y,z)$ denotes translation and $(R_x,R_y,R_z)$ denotes the axis-angle parameterization of rotation. This vector is then passed through a Fourier encoding:
\begin{align}
    E(p) = \big[\sin(2^k \pi u), \cos(2^k \pi u)\big]_{k=0}^{K-1}, \quad u \in \mathbb{R}^6,
\end{align}
which maps each dimension of $u$ to high-frequency features. The resulting embedding has dimension $e = 12K$.

We then train a classifier
\begin{align}
    f:\mathbb{R}^e \rightarrow \mathbb{R},
\end{align}
using a contrastive learning objective. For a batch containing positives $P^+$ and negatives $P^-$, we minimize the binary cross-entropy loss:
\begin{align}
    \mathcal{L}_{\text{feas}} = - \sum_{p \in P^+} &\log \sigma(f(E(p))) \nonumber \\
    &\ldots - \sum_{p \in P^-} \log \big(1 - \sigma(f(E(p)))\big),
\end{align}
where $\sigma(\cdot)$ denotes the sigmoid function.  

After training, the classifier produces a feasibility score
\begin{align}
    S_{\text{feas}}(p) = \sigma(f(E(p))),
\end{align}
which approximates the probability that pose $p$ corresponds to a stable grasp. This learned feasibility check generalizes beyond the discrete set of grasps generated by GPG and provides differentiable gradients, making it suitable for joint optimization with trajectory imitation and collision avoidance.

\subsubsection{Grasp Distance and Evaluation}

A key challenge of the JFTO framework is balancing the feasibility and consistency of a candidate robot grasp with those observed in demonstrations. To quantify the similarity between robot and human poses, we define a weighted distance on $\SE(3)$ that accounts for both translation and orientation. For two poses $p_1, p_2 \in \SE(3)$, let $t(p) \in \mathbb{R}^3$ denote its translation and $\phi(p) \in \mathbb{R}^3$ the axis--angle representation of its rotation. The distance is
\begin{align}
    D_{\SE(3)}(p_1, p_2) 
    =& \gamma \, \|t(p_1) - t(p_2)\|_2 \nonumber\\
    & \ldots + (1 - \gamma)\, \|\phi(p_1) - \phi(p_2)\|_1,
\end{align}
where $\gamma \in [0,1]$ balances the contributions of translation and orientation. The weight can be tuned by task; for example, orientation dominates in pouring tasks, while position is critical in pick-and-place tasks.

From the upstream pipeline, we extract a set of human grasps $H = \{h_i\}$, although not all are physically feasible due to differences between the human hand and the robot gripper. For a candidate robot grasp $p \in \SE(3)$, we measure similarity as the minimum distance from the demonstration set:
\begin{align}
    D_H(p) = \min_{h \in H} D_{\SE(3)}(p, h).
\end{align}

Given a robot joint configuration $q_0 \in \mathcal{J}$, we compute the corresponding gripper pose by forward kinematics,
\begin{align}
    p(q_0) = F(q_0) \in \SE(3).
\end{align}
This pose is then evaluated with two complementary criteria:
\begin{enumerate}
    \item \textbf{Feasibility:} A classifier $f:\mathbb{R}^e \to \mathbb{R}$ predicts grasp validity. The input to $f$ is a Fourier positional encoding $E:\SE(3)\to \mathbb{R}^e$ of the grasp pose. The output is squashed with a sigmoid $\sigma$, producing a probability-like score $\sigma(f(E(p(q_0))))$.
    \item \textbf{Grasp Demonstration Similarity:} The distance $D_H(p(q_0))$, which is small when the grasp is close to the one observed in demonstrations.
\end{enumerate}

The final grasp score is defined as
\begin{align}
    S_G(q_0) = \sigma(f(E(p(q_0)))) - \lambda\, D_H(p(q_0)),
\end{align}
where $\lambda > 0$ balances feasibility and similarity. High scores are assigned to grasp poses that are both executable (validated by the classifier) and close to human demonstrations (low $D_H$), encouraging the robot to select grasps that are robust and human-like.

\subsection{Collision Avoidance}\label{sec:COLN_METHOD}
While grasping and trajectory imitation ensure that the robot behaves like the demonstrator, safety requires avoiding collisions with the environment. To model the static scene, we fuse the background point clouds $B \subset \mathbb{R}^3$ obtained from the video into a dense reconstruction. From this cloud, we train a distance function $dist_B: \mathbb{R}^3 \rightarrow \mathbb{R}$, which outputs the estimated distance from any query point to the nearest obstacle surface. The differentiable field can be queried efficiently, making it suitable for gradient-based optimization.

For a given joint state $q_i \in \mathcal{J}$, we approximate the geometry of the robot arm by a finite set of surface points $Q(q_i) \subset \mathbb{R}^3$. These can be obtained either from mesh models sampled at fixed density or from bodypoints on the robot augmented with collision radii. A configuration $q_i$ is considered in collision if there exists $p \in Q(q_i)$ with $dist_B(p) < \ell$, where $\ell$ is a small safety margin (e.g., $0.01$\,m). To penalize unsafe motion over a rollout, we compute a collision score
\begin{align}
    S_C(q) = \frac{1}{T} \sum_{t=0}^{T} 
    \sum_{p \in Q(q_t)} \frac{1}{\ell} \, \max(0,\ell - dist_B(p)),
\end{align}
Here, $ \max(0,\cdot)$ ensures that only points within the safety margin contribute, while scaling by $1/\ell$ normalizes the penalty. Furthermore, $S_C(q)$ remains close to zero for safe trajectories but increases as the robot approaches or intersects obstacles, creating strong gradients that steer optimization toward collision-free solutions. This formulation also allows ``near-miss behaviors'' (where the robot comes close to the environment) to be penalized less severely than actual interpenetration, encouraging motions that are safe yet not overly conservative.
\begin{algorithm}[t]
\caption{Joint Flow Trajectory Optimization}
\label{alg:motion_planning}
\small
\begin{algorithmic}[1]
    \STATE Trajectory $\xi = \{q_t\}_{t=0}^{T}$, optimization weighting hyperparameters $\alpha$, $\beta$, $\gamma$
    \STATE Set optimizer $\gets$ Adam($\xi$)
    \FOR{$i = 1$ to $N$ steps}
        \STATE Compute $S_T(\xi)$ (See \cref{sec:TRAJ_METHOD})
        \STATE Compute $S_G(q_0)$ (See \cref{sec:GRASP_METHOD})
        \STATE Compute $S_C(\xi)$ (See \cref{sec:COLN_METHOD})
        \STATE loss $\gets -[\alpha S_T(\xi) + \beta S_G(q_0) + \gamma S_C(\xi)]$
        \STATE Minimise gradients of loss w.r.t. $\{q_t\}_{t=0}^{T}$
        \STATE Update $\xi$ with optimizer
    \ENDFOR
    \STATE \textbf{return} $\xi$.
\end{algorithmic}
\end{algorithm}

\subsection{Algorithm Outline}\label{sec:ALGO_METHOD}

An important property of the JFTO formulation is that each component of the objective, trajectory likelihood $S_T$, grasp score $S_G$, and collision penalty $S_C$, is fully differentiable. This allows us to directly optimize the joint trajectory $q$ using gradient-based methods. The procedure is summarized in Algorithm~\ref{alg:motion_planning}. This algorithm can be parallelized on the GPU where we optimize a large batch of trajectories. This improves robustness against poor local minima. The candidate with the highest final score is selected.

\begin{figure*}[t]
    \centering
    \includegraphics[width=\textwidth]{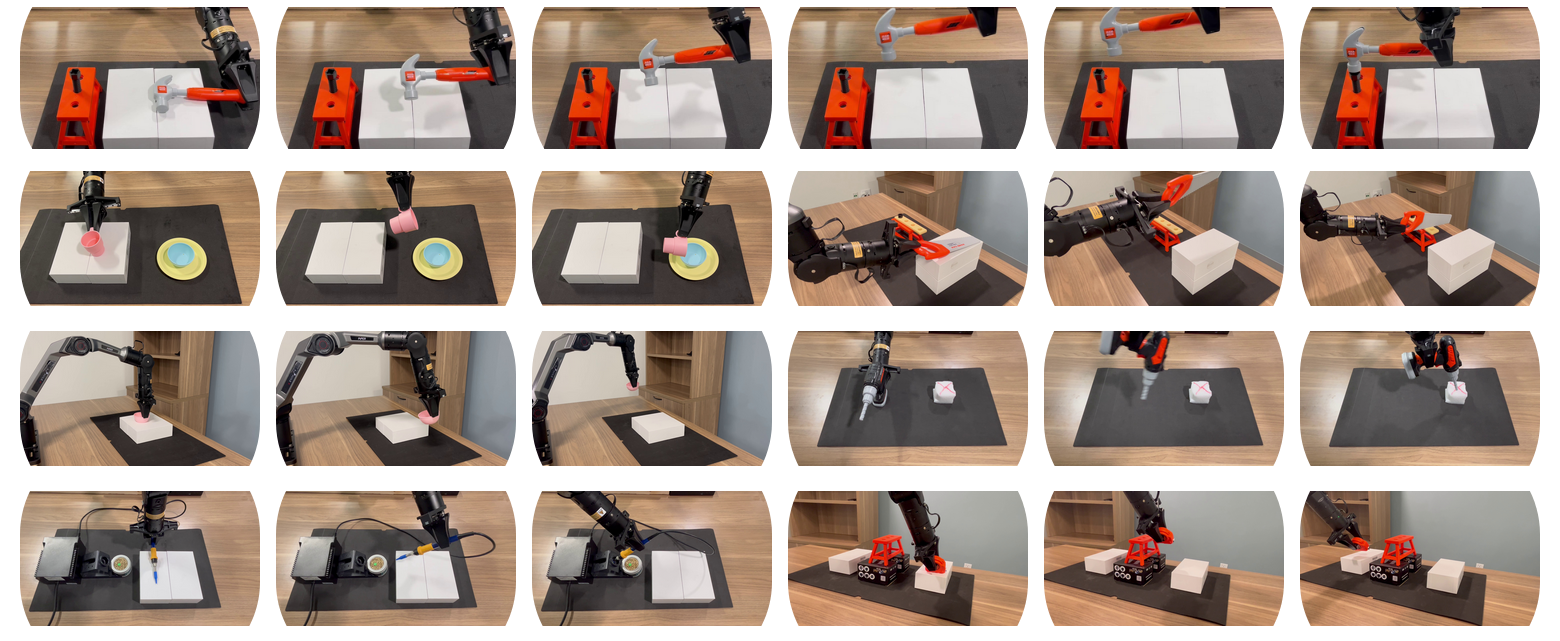}
    \caption{We present the experimental results of the $7$ real-world tasks. The first row corresponds to Task \ref{task:hammer}, second row first three corresponds to Task \ref{task:cup}, last three corresponds to Task \ref{task:saw}, third row first three corresponds to Task \ref{task:figure8}, last three corresponds to Task \ref{task:drill}, last row first three corresponds to Task \ref{task:solder}, last three corresponds to Task \ref{task:metertape}.}
    \label{fig:experimental_results}
    \vspace{-1.75em}
\end{figure*}

\section{Empirical Evaluations}\label{sec:EXPERIMENT}

We evaluate the proposed \emph{Joint Flow Trajectory Optimization} method in simulation and on a real robot. Our study asks: \textbf{(i)} Does joint optimization select grasps that are feasible \emph{and} conducive to faithful object-trajectory imitation? \textbf{(ii)} How does joint optimization compare to a sequential pipeline that optimizes the grasp then trajectory? \textbf{(iii)} Does modeling demonstration \emph{multi-modality} through flow matching improve imitation relative to distance-based objectives?

\textbf{Hardware and Platform:} Real-world trials are carried out on a \emph{Agile-X Piper} 6-DoF arm with a parallel gripper. The limited DoF and small work envelope increase the likelihood of joint-limit and reachability conflicts, providing a challenging setting for grasp feasibility, collision avoidance, and faithful trajectory execution.

\textbf{Tasks and Demonstrations:} We construct a seven-task benchmark spanning tool use, pouring, cutting, drawing-like motions, and obstacle-aware transport:

\noindent\task{task:hammer}{Hammer Hitting Nail}: pick up a hammer and strike a nail.

\noindent\task{task:cup}{Cup Pouring Water}: grasp a cup and pour water into a bowl on the left.

\noindent\task{task:saw}{Saw Cutting Wood}: lift a saw from the base, reorient upright and perform a cutting motion.

\noindent\task{task:figure8}{Drawing Figure 8}: trace a figure-eight with a held bowl.

\noindent\task{task:drill}{Drilling Foam}: grasp a toy drill and drill into foam.

\noindent\task{task:solder}{Solder Putting Back}: return a soldering tool to its station.

\noindent\task{task:metertape}{Shifting Meter Tape}: transport a tape across the workspace. We introduce additional obstacles in inference.

\begin{figure}[t]
    \centering
    \includegraphics[width=0.8\linewidth]{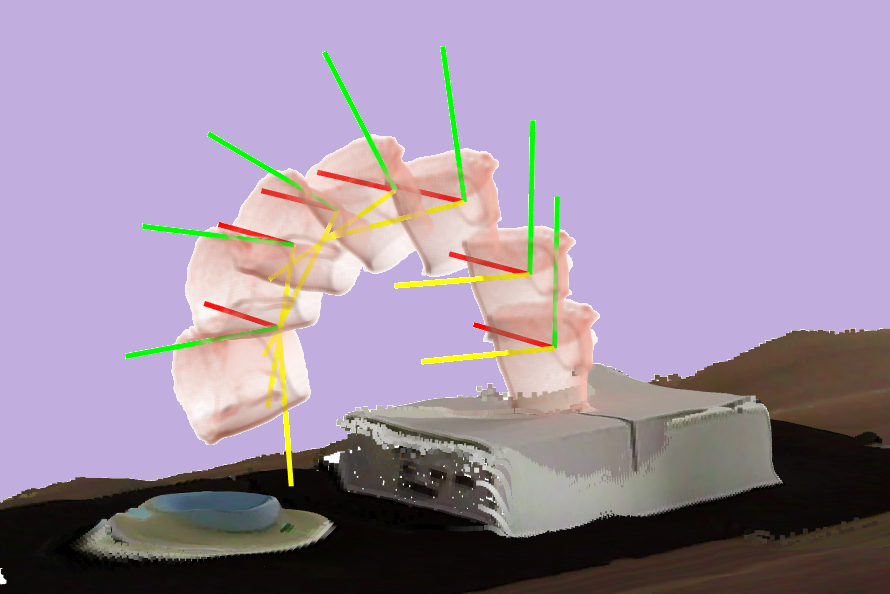}
    \caption{We applied the upstream processing method over each key frames in the video to get a trajectory of object poses. Here, we observe the object trajectory of Task \ref{task:cup}.}
    \label{fig:trajectory-demo-vis}
\end{figure}

Human demonstrations are processed using the pipeline described in \cref{sec:DATA_COLC}: stereo DUSt3R~\cite{Wang_2024_CVPR} for 3D reconstruction with metric scaling, SAM2~\cite{ravi2024sam2} for hand–object segmentation, HaMeR~\cite{pavlakos2024reconstructing} for hand pose estimation, and ICP-based tracking to recover object trajectories $\{\hat{x}_t\}_{t=0}^{T}$. A fused static background is additionally constructed for distance-field collision checking. \Cref{fig:trajectory-demo-vis} illustrates an example trajectory from Task~\ref{task:cup}, where the reconstructed 3D motion of the cup can be visually traced.

The execution of the tasks is illustrated in \Cref{fig:experimental_results}. The set of evaluated tasks span a diverse set of motions, including tool use, pouring, cutting, and obstacle-aware transport. Together, they provide a representative and challenging benchmark for assessing whether the proposed method can be generalized across different manipulation styles.

\begin{figure*}[t]
    \centering
    \begin{minipage}{0.7\textwidth}
        \centering
        \captionof{table}{Comparison of trajectory similarity between joint flow trajectory optimization and sequential optimization. Joint optimization generally yields lower rotational error and higher average log density, indicating better alignment with human demonstrations.}
        \label{tab:joint_vs_sequential}
        \begin{adjustbox}{width=\linewidth}
        \begin{tabular}{lccccccc}
            \toprule
            Method & Hammer & Cup & Saw & Figure 8 & Drill & Solder & Meter Tape \\
            \midrule
            \multicolumn{8}{l}{\textbf{Joint Optimization}} \\
            \hline
            Avg Dist Diff $\downarrow$ & 0.0108 & 0.0150 & 0.0137 & 0.0116 & 0.0181 & 0.0169 & 0.0270 \\
            Avg Rot Diff $\downarrow$ & 0.2572 & 0.3180 & 0.2160 & 0.1787 & 0.4543 & 0.3647 & 0.1686 \\
            Avg Log Density $\uparrow$ & 7.0973 & 6.2451 & 7.2066 & 7.4351 & 7.0655 & 6.0329 & 6.8647 \\
            \midrule
            \multicolumn{8}{l}{\textbf{Sequential Optimization}} \\
            \hline
            Avg Dist Diff $\downarrow$ & 0.0171 & 0.0110 & 0.0123 & 0.0080 & 0.0087 & 0.0296 & 0.0253 \\
            Avg Rot Diff  $\downarrow$ & 0.5748 & 0.4075 & 0.4263 & 0.2758 & 0.3968 & 0.8603 & 0.1518 \\
            Avg Log Density $\uparrow$ & 1.4823 &-2.7326 & 1.6412 & 6.5238 & 5.7580 &-1.6030 & 5.7239 \\
            \bottomrule
        \end{tabular}
        \end{adjustbox}
    \end{minipage}\hspace{0.5em}
    \begin{minipage}{0.24\textwidth}
        \centering
        \includegraphics[width=0.43\linewidth]{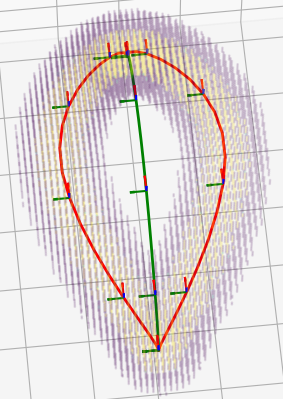}
        \includegraphics[width=0.53\linewidth]{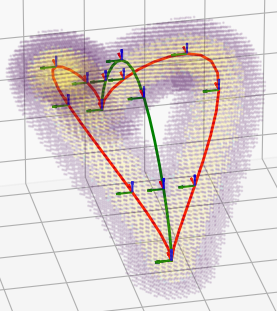}
        \captionof{figure}{Visualization of trajectory distributions. Yellow/purple indicates high/low density regions. Flow-based optimization converges to one of the demonstrated red trajectories. Distance-based optimization collapses toward the mean (in green).}
        \label{fig:multi-modality}
    \end{minipage}
    \vspace{-1.75em}
\end{figure*}

\subsection{Evaluation Metric}

The JFTO framework produces a sequence of joint states, which we map through forward kinematics to obtain the end-effector trajectory. Using the first end-effector pose together with the initial object pose, we compute the rigid transform between the gripper and the object. This allows us to derive the executed object trajectory induced by the rollout.

To measure how closely the executed trajectory follows the human demonstration, we report both translational and rotational errors averaged across timesteps:
\begin{align}
    \Delta dist_{avg} &= \frac{1}{T} \sum_{t=0}^{T} \big\lVert \mathbf{p}_{t}^{demo} - \mathbf{p}_{t}^{exec} \big\rVert_2, 
\end{align}
\begin{align}
    \Delta rot_{avg} &= \frac{1}{T} \sum_{t=0}^{T} R\!\left(\mathbf{r}_{t}^{demo}, \mathbf{r}_{t}^{exec}\right),
\end{align}

where $\mathbf{p}_{t}^{demo}, \mathbf{r}_{t}^{demo}$ are the position and rotation of the pose of the object demonstrated in timestep $t$, and $\mathbf{p}_{t}^{exec}, \mathbf{r}_{t}^{exec}$ are those of the executed trajectory. The function $R(\cdot,\cdot)$ denotes the angular distance between two orientations.

In addition to these geometric errors, we also compute \emph{average log density} of the poses of the executed object under the demonstration-informed density model introduced in \cref{sec:TRAJ_METHOD}. This provides a probabilistic measure of how well the rollout aligns with the distribution of human demonstrations.

\subsection{Empirical Performance and Hyperparameter Choices}
Our proposed JFTO framework is relatively insensitive to hyper-parameters, which trade off each of the desired loss terms. In our evaluations, we choose hyperparameters to balance grasp feasibility, trajectory imitation, and collision avoidance. Collisions are \emph{strictly} penalized by setting $\gamma=1$, reflecting their critical impact on execution safety. For the trade-off between the grasp and the trajectory terms, $\alpha=\beta=0.5$ yields balanced performance but occasionally leads to unstable grasps under some random seeds. Raising the grasp weight to $\beta=0.8$ consistently improves reliability, securing successful grasps without significantly degrading trajectory quality. Although selected empirically, these settings align with the practical intuition that grasp stability should be prioritized for robust real-world execution.  

\subsection{Joint vs.\ Sequential Optimization}
We compare our joint flow trajectory optimization (JTFO) against a \emph{sequential} baseline, in which the grasp is first optimized to match the demonstration, then held fixed while a second stage optimizes the object trajectory. In other words, the baseline selects a feasible, demonstration-like grasp and subsequently rolls out a trajectory that minimizes deviation from the demos.

JTFO successfully executes all seven tasks (qualitative results in \cref{fig:experimental_results}); quantitative metrics are reported in \cref{tab:joint_vs_sequential}. Both methods achieve similarly low \emph{positional} errors, indicating that objects are generally delivered to the correct locations. However, the sequential baseline incurs substantially higher \emph{rotational} errors and markedly lower log densities on tasks such as \emph{Hammer}, \emph{Cup}, \emph{Saw}, and \emph{Solder}. This suggests that, although the baseline can identify a grasp near the demonstrated one, it often fails to maintain the correct object orientation during execution. In contrast, JTFO reasons about execution feasibility \emph{during} grasp selection. Although this can slightly reduce initial grasp similarity, it yields object poses that remain consistent over time. Consequently, JTFO produces higher-likelihood trajectories that more faithfully reproduce the demonstrated motions.

\subsection{Multi-Modality: Why Flow Matching}\label{sec:multi_modal}

Many real-world manipulation tasks admit multiple valid strategies for achieving the same goal. For instance, an object may be moved around an obstacle either to the left or to the right. In such cases, the demonstration distribution is inherently \emph{multi-modal}. Traditional distance-based objectives, such as minimizing the Euclidean distance to all demonstrations, cannot capture this structure. They encourage trajectories to converge toward an ``average'' solution that lies between modes-often producing motions that are physically infeasible or semantically incorrect.

Flow matching addresses this limitation by explicitly modeling the density of the demonstration distribution. Rather than collapsing distinct behaviors into a single mean, the learned velocity field guides optimization toward one of the high-density regions in $\SE(3)$ that correspond to observed demonstrations. The mode being selected depends on the initialization of the optimization, but in all cases the resulting trajectory is consistent with at least one valid human strategy. This property is illustrated in \cref{fig:multi-modality}. The yellow regions represent areas of high demonstration density. Gradient descent on the learned flow drives sampled waypoints into these regions, producing trajectories that align with observed demonstrations (red paths). In contrast, a Euclidean distance objective yields an intermediate solution (green path), corresponding to an unrealistic average trajectory not observed in the data. By preserving multi-modality, flow matching ensures that robots reproduce \emph{plausible human-like strategies}, rather than collapsing behaviors into averages.

\begin{figure}[t]
    \centering
    \includegraphics[width=\linewidth]{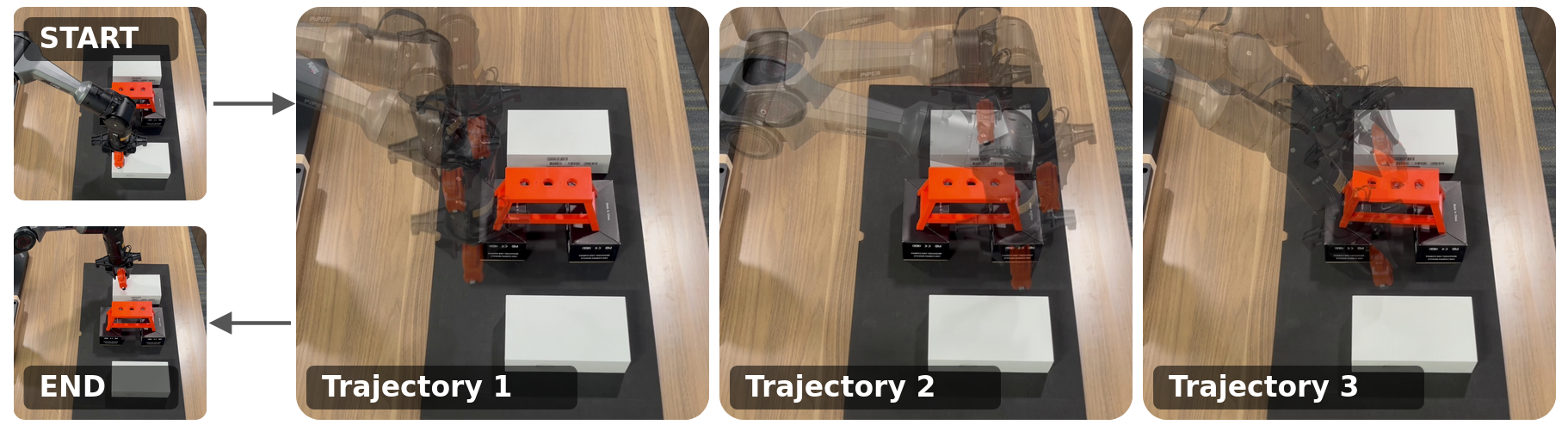}
    \caption{Given multi-modal trajectory distributions (avoiding obstacles), the flow matching model yields either Trajectory~1 or Trajectory~2, depending on initialization, both of which follow a demonstrated strategy. In contrast, the distance-based model mode-seeks and produces Trajectory~3, cutting through the obstacle in a way not consistent with any human demonstration.}
    \label{fig:multi-modal-demo}
\end{figure}

Here, we extend Task \ref{task:metertape} (moving meter tape) by collecting multiple demonstrations around an obstacle: Some routes placed the tape in front, while others carried it behind. Leveraging the flexibility of our JFTO framework with flow matching, the robot reliably reproduced one of these demonstrated strategies--passing in front or behind--depending on initialization (trajectories~1 and~2 in \cref{fig:multi-modal-demo}). Both outcomes correspond to faithful imitation. In contrast, applying a naive distance-based metric over the entire set of demonstrations yields an averaged path that attempts to move directly through the obstacle, bending upward to compensate (trajectory~3 in \cref{fig:multi-modal-demo}). This trajectory does not resemble any of the demonstrations.

\section{Conclusion and Future Work}
In this paper, we address generating feasible grasp poses and motion trajectories from video-based LfD, ensuring that robot grasps and object motions remain consistent with human demonstrations despite kinesthetic infeasibility. We presented the Joint Flow Trajectory Optimization (JFTO) framework that integrates grasp selection and trajectory imitation within a single optimization process. By leveraging flow matching as a density estimator, our method naturally handles multi-modal human demonstrations, avoiding unrealistic averaging effects and guiding the robot toward feasible, human-like behaviors. Experiments in both simulation and on a low-cost 6-DoF robotic arm demonstrated that joint optimization consistently outperforms sequential baselines, achieving lower rotational error and higher trajectory likelihood. These results highlight the practicality and robustness of our approach for transferring human demonstrations to resource-constrained robotic platforms. As many human tasks require using both hands to interact with objects, a clear line of future inquiry is to extend JFTO to setups where bimanual manipulation is involved, where the human video demonstration shows the human using both hands.

\bibliographystyle{ieeetr} 
\bibliography{bib}
\end{document}